\documentclass[sigconf]{acmart}
\usepackage{colortbl}
\usepackage{pgfplots}
\usepackage{pgfplotstable}

\usepackage{booktabs} 

\usepackage{graphicx}
\usepackage{epsfig}
\usepackage{amssymb}
\usepackage{amsmath}
\usepackage{helvet}
\usepackage{courier}

\usepackage{etoolbox}
\usepackage{color}
\usepackage{epsfig,amsfonts,latexsym}
\usepackage{float}
\usepackage{graphicx}
\usepackage{multirow}
\usepackage{caption}
\usepackage{subcaption}
\usepackage{enumitem}
\usepackage{xcolor}
\floatstyle{plaintop}
\restylefloat{table}
\restylefloat{figure}
\usepackage{color}
\usepackage{algorithm,algorithmic}
\usepackage{caption}
\usepackage{blindtext}

\usepackage[framemethod=TikZ]{mdframed}
\usepackage{lipsum}
\mdfdefinestyle{MyFrame}{%
    linecolor=blue,
    outerlinewidth=2pt,
    roundcorner=20pt,
    innertopmargin=\baselineskip,
    innerbottommargin=\baselineskip,
    innerrightmargin=20pt,
    innerleftmargin=20pt,
    backgroundcolor=white}

\newcount\Comments  
\usepackage{color}
\definecolor{darkgreen}{rgb}{0,0.5,0}
\newcommand{\kibitz}[2]{\ifnum\Comments=1\textcolor{#1}{#2}\fi}

\usepackage{hyperref}
\hypersetup{colorlinks,
citecolor=blue,
linkcolor=blue,
urlcolor=black
}
\setlist[itemize]{noitemsep, topsep=0pt}

\setcopyright{rightsretained}

\pgfplotstableset{
    /color cells/min/.initial=0,
    /color cells/max/.initial=1000,
    /color cells/textcolor/.initial=,
    color cells/.code={%
        \pgfqkeys{/color cells}{#1}%
        \pgfkeysalso{%
            postproc cell content/.code={%
                \begingroup
                \pgfkeysgetvalue{/pgfplots/table/@preprocessed cell content}\value
                \ifx\value\empty
                    \endgroup
                \else
                \pgfmathfloatparsenumber{\value}%
                \pgfmathfloattofixed{\pgfmathresult}%
                \let\value=\pgfmathresult
                \pgfplotscolormapaccess
                    [\pgfkeysvalueof{/color cells/min}:\pgfkeysvalueof{/color cells/max}]
                    {\value}
                    {\pgfkeysvalueof{/pgfplots/colormap name}}%
                \pgfkeysgetvalue{/pgfplots/table/@cell content}\typesetvalue
                \pgfkeysgetvalue{/color cells/textcolor}\textcolorvalue
                \toks0=\expandafter{\typesetvalue}%
                \xdef\temp{%
                    \noexpand\pgfkeysalso{%
                        @cell content={%
                            \noexpand\cellcolor[rgb]{\pgfmathresult}%
                            \noexpand\definecolor{mapped color}{rgb}{\pgfmathresult}%
                            \ifx\textcolorvalue\empty
                            \else
                                \noexpand\color{\textcolorvalue}%
                            \fi
                            \the\toks0 %
                        }%
                    }%
                }%
                \endgroup
                \temp
                \fi
            }%
        }%
    }
}

\begin{document}
\title{A relevance-scalability-interpretability tradeoff with temporally evolving behavioral personas}
\titlenote{This work was achieved while both authors were employed by Technicolor, Los Altos, CA. }

\author{Snigdha Panigrahi}
\affiliation{%
  \institution{Stanford University}
  \city{Stanford} 
  \state{CA} 
  \postcode{94305}
}
\email{snigdha@stanford.edu}

\author{Nadia Fawaz}
\affiliation{%
  \institution{LinkedIn}
  \city{Sunnyvale} 
  \state{CA} 
}
\email{Nfawaz@linkedin.com}

\renewcommand{\shortauthors}{S. Panigrahi et al.}

\begin{abstract}
The current work characterizes the users of a VoD streaming space through user-personas based on a \textbf{tenure timeline} and \textbf{temporal} behavioral features in the absence of explicit user profiles. A combination of tenure timeline and temporal characteristics caters to business needs of understanding the evolution and phases of user behavior as their accounts age. The personas constructed in this work successfully represent both dominant and niche characterizations while providing insightful maturation of user behavior in the system. 
The two major highlights of our personas are demonstration of \textit{stability} along tenure timelines on a population level, while exhibiting interesting migrations between labels on an individual granularity and clear interpretability of user labels. Finally, we show a trade-off between an indispensable trio of guarantees, \textit{relevance-scalability-interpretability} by using summary information from personas in a CTR (Click through rate) predictive model. The proposed method of uncovering latent personas, consequent insights from these and application of  information from personas to predictive models are broadly applicable to other streaming based products. \end{abstract}

\maketitle

\section{Introduction}
\label{intro}
User segmentation, the idea of dividing a market up into homogeneous segments and targeting each group with a distinct product or message is a basic tool to model similar consumers. This is explored in diverse sectors like finance \cite{li2010credit}, health \cite{gummerus2004customer}, telecommunications \cite{bayer2010customer} etc. and through focus on different behavioral aspects in \cite{agrawal1996quest}, \cite{apte2002business}, \cite{chen2005mining}. The current work adopts a latent parametric \textit{mixture model} approach to construct segments of homogeneous consumers called \textit{user personas} for VoD services from raw transactional logs, using a tenure timeline and temporal behavioral features. 
Examples of such services in the VoD space include \textit{itune, googleplay, vudu, FandangoNow}, etc; where users pay per piece of content they watch. This is in contrast with subscription based services, where users pay a monthly subscription, such as netflix, hulu plus, amazon video etc. The work provides explicit user characterizations on \textbf{spending behavior}, \textbf{content preference} and \textbf{transactional habits} with the main contributions as presented below:
\begin{itemize}[noitemsep,topsep=0pt]
\item align user transaction timelines on a \textbf{tenure basis} at a \textbf{monthly} granularity, a novel choice for a timeline of comparison, in place of the conventional calendar timeline 
\item construct temporal feature vectors from transaction logs, that are aggregates of transactions over a month along tenure timeline; such features represent the \textbf{evolving behavioral} consumer traits
\item capture both \textbf{dominant} and \textbf{niche} segments of population and provide highly interpretable user labels
\item capture stable latent structure on a population level, even as individual profiles keep transforming with age; personas maintain a \textbf{consistent} clustering over time while accurately explaining the changes on an individual level
\item represent insights on inter-relations between behavioral characteristics as layers within user profiles.
\end{itemize}
Such a construction of temporally evolving personas with new insights into behavioral characteristics is the \textit{first of its kind} in the streaming space, to the best of our knowledge.
 
A line of prior works \cite{allenby1998marketing}, \cite{besanko2003competitive}, \cite{bhatnagar2004latent}, \cite{wu2005research} \cite{wedel2012market} has explored characterization of consumers; another independent set of works has contributed to methods on personalized recommendations \cite{ansari2000internet}, \cite{sarwar2001item}, \cite{schafer2007collaborative}, \cite{pazzani2007content}, \cite{hu2008collaborative}, \cite{ekstrand2011collaborative}, \cite{koren2009matrix}. Our work concludes with a unification of these two important goals to demonstrate the utility of \textit{user personas}. In particular, we illustrate an application of persona based features in CTR (Click through Rate) predictions. We show that a model based on the constructed personas achieves a 3 criteria \textbf{relevance - scalability - interpretability} tradeoff, when compared against models that do not include persona information. We show a substantive gain in computational cost through the use of lower dimensional persona features in the form of soft or hard clustering information. This gain occurs with retaining clarity in the interpretation of feature space (as opposed to random projections onto lower dimensional spaces) and does not compromise with predictive ability. The CTR model we describe is interesting in its own right as we use persona features in a logistic model trained per item to capture item specific variability. The use of persona information can also aid in preserving anonymity of individual users as well as of individual transactions. 
We supplement the CTR model with a discussion on other commonly used collaborative filtering models that can potentially achieve a similar trade-off.

Our methods are by no means limited to the VoD space. They can be extended to lend similar insights and achieve similar benefits for other product based services. Modeling latent structure from raw transactional data can overcome the curse of dimensionality through an efficient reduction in regression size, while maintaining predictive power and interpretability of feature space.

\subsection{Related works}
\label{rel:works}
Consumer segmentation, driven by the intuition that predictive models of customer behavior based on groups of similar customers outperform a single aggregate model, dates back to \cite{smith1956product}. A segmented predictive model in \cite{allenby1998marketing},\cite{besanko2003competitive},\cite{wedel2012market} can be refined further to an individual level, trained per customer. In doing so, we gain a reduced bias in creating increasingly more homogeneous customer groups at the cost of increased variance in estimation as we consider progressively more refined segments containing fewer customers. Thus, there is a classic bias-variance trade-off which is effectively dealt by integrating customer segmentation into such predictive models, termed as \textit{segmented models} in \cite{jiang2006segmenting}. In this work, we advocate the use of features based on user personas not only for improvement of predictive power but, as a \textit{meaningful, lower dimensional, summary} space that can be used to achieve scalability in regression models and facilitate storage for future debugging. 

Various techniques of segmenting consumers include neural net models in \cite{vellido1999segmentation} \cite{boone2002retail}, latent probabilistic models in \cite{dias2007latent},\cite{jedidi1997finite}, combinatorial optimization based grouping models in \cite{jiang2009improving}.  We offer in this work a multinomial latent mixture model analysis with both soft and hard clustering values as outputs, employing the classic EM algorithm in \cite{dempster1977maximum} to estimate the mixing proportions and distribution parameters for building user personas. Most part of the raw data-logs consists of count features for which a multinomial model seems a natural choice; except for the spending amounts which we choose to implement the K-means clustering which gives similar results as as the more commonly used parametric Gaussian mixture model \cite{friedman1997image}. In comparison to prior art, our goal here goes beyond discovering latency. That is, we want labels that can directly render business insights as opposed to non-interpretable clusters. 

One of the key features of our personas is that they exhibit stability on a population level even as migrations on an individual level are constantly taking place along the chosen time granularity. Clusters not shifting dramatically from one time-step to the next is also explored in \cite{chakrabarti2006evolutionary} and the stability of clusters finds similarity in equilibrium of average network properties in \cite{kossinets2006empirical}. The personas possess a natural divisive structure as opposed to imposed hierarchy via clusterings discussed in \cite{guha1998cure}, \cite{bandyopadhyay2003energy} amongst many others. 

\subsection{VoD Data set}
\label{data}
Dataset considered in this work consists of transaction logs of a subset of $\bf 730,000$ anonymous users from a large-scale streaming VoD service across a time span of $\bf 16$ months from January 2014 to April 2015, with over $\bf 2$ million transactions. 
Each record in the transaction logs consists of a unique user-id, a unique time-stamp, a unique content-id, the type of transaction--rentals/ purchases, a net price giving the cost of each transaction, and content meta-data such as genres, release year, MPAA ratings corresponding to each transaction. \cite{trouleaujust} analyzed a processed user-interactional part of this data set, consisting of $3488$ users and $26404$ viewing sessions, to model binge watching behavior for VoD services; we consider the full set of users in our analysis and focus on the transactional data base instead. 

We present few summary statistics based on the transactional data-base; these preliminary statistics and observations lead to the belief that there is latent structure in the users consumption patterns and guide the pre-processing stage to construct features from raw transaction logs. Note that the characterizations of user behavior discovered as latent structure from raw logs in this work can be viewed as more precise and refined summaries. 
The transactions break up into two types- rentals and purchases with $\bf 88\%$  rentals and $\bf 12\%$. The price categories of rentals vary from $\bf 0-5\$$ with higher price categories falling in the $\bf 3-5\$$ range. The purchases range as high as $\bf 25\$$, mainly for new movies and tv series. The purchases greater than $\bf 10\$$ in value are considered as higher end transactions. Most transactions occur in the lower price categories of both types of transactions with only $\bf 10\%$ of consumers transacting in the higher price ranges. A transactional perspective of the content catalogue is observed through segregation of transactions into $\bf 15\%$ TV shows and $\bf 85\%$ movies, with the movie \textit{Frozen} being the most consumed content in the catalogue. The dominant genres in the transactions are \textit{ Drama ($\bf 18\%$), Comedy ($\bf 10\%$), Action ($\bf 10\%$), Family ($\bf 9\%$), Animation ($\bf 7\%$), Thriller ($\bf 6\%$), Biography ($\bf 5\%$), Sci-fi ($\bf 4\%$), Crime ($\bf 4\%$)} etc, with the crucial observation that while some users ($\bf 20\%$) tend to prefer more family-friendly content (Family, Animation, Super-hero). Other segments ($\bf 80\%$) consume genres such as drama, horror, comedy etc. The time of transactions is seen mostly to range between evenings and nights, evenly split between weekdays (Monday-Friday) and weekends.
As part of the pre-processing of raw logs, barely active users (spend less than $\bf 1$ dollar in a certain month of activity) and one-time deal hunters (transact only once and never return) were filtered out to prevent cluster centers being pulled to $\bf 0$. Summarized information is subsequently uncovered from the data as cluster centers and cluster sizes, which preserves anonymity of individual users while not giving information on any particular transaction.
\section{Construction of user personas}
\label{userpersonas}
We offer a behavioral segmentation based on spending traits, content preferences and transactional habits of users with interest in above characterizations stemming from intuition, domain knowledge and business goals. Before, we delve into the model for describing these characterizations, we discuss the timeline and granularity of comparison and a brief description of features (constructed as aggregates over a month of transactions, binned appropriately), crucial in excavating meaningful latent structure in data.
\subsection{Timeline of comparison}
Transaction logs consist of time-series data. We make a careful choice as to how the timelines of different users are compared with regard to the following 2 aspects:\\
\textbf{Temporal alignment of user timelines}: User timelines can be aligned on a calendar basis or on a tenure basis. In the calendar basis, transactions of different users happening at the same calendar dates, for instance in January 2014, are compared against each other. Aligning timelines according to a calendar basis allows to detect seasonalities(holidays, end-of-year movie releases), and effects of specific events happening at a particular date (TV-show new episode/season release or end). On the other hand, in the tenure basis, the first transaction of a given user defines the birth of the user timeline, and transactions of different users are compared when they happen at the same age of the user in the system. For instance, if user A made his first transaction on January, 15th 2014 and user B made his first transaction on April, 10th 2014, comparisons would be drawn for their first month of transactions between Jan. 15th-Feb 14th 2014 for user A and April 10th-May 9th for user B. Aligning timelines on a tenure basis allows observations on how users age in the system and helps in understanding behavioral phases and in predicting churn.\\
\textbf{Temporal granularity}: Timestamps in transaction logs can be specified up to seconds or even milliseconds. When building features based on time-series, the question arises as to the granularity at which events should be grouped to devise the desired features. Transactions can be aggregated at a monthly/ weekly/ daily/hourly granularity. For instance, to compute a count feature at the monthly granularity, transactions happening within the same 30 day period will be aggregated. The granularity level affects the detection of behavioral patterns and cycles.
In this work, user timelines are aligned on a \textbf{tenure basis}, and events are considered at a \textbf{monthly (30 days) granularity}.
The first transaction of a user marks the beginning of its timeline, and user's transaction history is divided into successive periods of 30 days each.
Our choice of a monthly granularity is guided by elementary analysis of the transaction logs which show unstable structures with weekly granularity-- \textit{Weekly logs were too short a period to capture behavioral patterns}--, and a flat structure at a quarterly granularity--\textit{Quarterly logs were too long to capture the dynamism in user labels due to an over-cumulation effect of data}.
Our choice of a tenure basis was motivated by the business need to understand the evolution and phases of user behavior along their transaction histories; this helps model their dynamic behavior, predict loss of interest in system, predict lifetimes etc.
\subsection{Aggregate feature space}
The features used in construction of personas are aggregates of transactions at monthly granularity, binned into categories. The choices of binning, arising from a combination of summary knowledge of data and domain information, lead to the below features.\\
\textbf{Monthly Expenditure (ME)} characterizes spending behavior: Each feature is the total net amount spent in one month by a user in either a rental/purchase transaction type and a given price category (5 for rentals, 8 for purchases).\\
\textbf{Transaction frequency (TF)} characterizes economic behavior: Features are transaction counts binned into 2 price categories in rentals and 4 categories in purchases.\\
\textbf{Dominant genres (DG)} indicates content preference: Features are monthly counts of transactions in 15 most popular genres: \textit{ Drama, Comedy, Action, Family, Animation, Thriller, Biography, Sci-Fi, Crime, Super Hero, Comedy-Drama, Fantasy, Horror, Romance, Kids, Miscellaneous.}\\
\textbf{Content recency (CR)} indicates freshness preference: Features are counts binned into ranges of content release year: \textit{Old: $<1990$, Nostalgia:  $1990-2000$, Not New: $2000-2010$, Recent:$2010-2013$ and Latest: $2014-2015$}.\\
\textbf{Time \& day of transaction (TDT)} gives transacting habits: Time-stamps of transactions are processed to generate the day of week and time of transaction as per the geographic region of the user, then counts are binned into weekdays or weekends and 4 time slots: 
\textit{10 AM-5 PM (Office Hours), 5PM-10 PM (evening and night), 10PM-5AM (late night).} 

\subsection{A mixture model for latent characteristics}
\label{mmm}
To fix notations for this section, we have a $n \times d$ feature matrix \[X^T=(x_1,x_2,\cdots,x_n),\] with $x_i\in \mathbb{R}^d$ representing the feature vector of user $i$ in a sample of $n$ users. We propose a parametric approach, a mixed multinomial model \noindent\textbf{MMM}, implementations of which are seen in  \cite{dunson2000bayesian, chiu2001robust,  vermunt2002latent, hedeker2003mixed, erosheva2004mixed} to describe user labels based on count data. The choice of a multinomial distribution is motivated from the very fact that it is an extremely natural model for count feature vectors. The iterative EM algorithm applied to estimate the mixing proportions and the parameters in mixed multinomial distribution, is in itself a very powerful mechanism, with one of its many merits being the ability to deal with missing features. A more common model is the Gaussian mixture model GMM, applied in \cite{reynolds1995robust, permuter2003gaussian}.\\
A MMM is based on the assumption that the rows of $X$ are independent draws from a multinomial model, that is
$x_i \sim MN(d,\theta_{Z_i})$ where $Z_i$ being a latent variable taking values $j\in [K]$, K- representing the number of clusters; independent of $X_i$. We have a hierarchical structured model as 
$$Z_i \stackrel{iid}{\sim} MN(1,\pi); \; X_i\lvert Z_i \stackrel{ind}{=} z \sim MN(d, \theta_z),$$ with  $\pi=(\pi_1,\cdots, \pi_K)$ representing mixing probabilities for the K clusters and $\theta_j =(\theta_{j,1},\cdots, \theta_{j,d});\; j\in[K]$ respresenting the multinomial parameters in density given latent factor $Z=j$.

 The mixing probabilities $\pi$ and parameters of mixture model $\theta_j, j\in[K]$ are estimated using an EM algorithm as proposed in \cite{dempster1977maximum}. We outline the E and M steps for the $(t)$-th iteration of the algorithm for the MMM based on iterates $\pi^{(t)}$ and $\theta^{(t)}$--\\
\textbf{E-step}: computes the posterior probabilities given estimates of parameters $\pi$ and $\theta_{Z_j}$ of the $t$-th iteration, that is
\begin{equation*}
\begin{aligned}
\tau_{i,z}^{(t)}&=\mathbb{P}(Z_i=z\lvert x_i;\pi^{(t)},\theta^{(t)}) \\
&=\dfrac{\mathbb{P}(X=x_i\lvert Z_i=z; \;\theta^{(t)})\pi^{(t)}_z}{\sum_{j=1}^{k}\mathbb{P}(X=x_i\lvert Z_i= j;\theta^{(t)})\pi^{(t)}_{j}}.
\end{aligned}
\end{equation*}
where $\mathbb{P}(X=x_i\lvert Z_i= j;\theta^{(t)}) = \Pi_{v=1}^{d}\theta_{j,d}^{x_{i,d}}$.\\
\textbf{M-step}: maximizes the Expected Complete Log Likelihood (ECLL) to refine estimates of parameters $\pi$ and $\theta_j$ for $j\in [K]$;
$$\theta^{(t+1)},\pi^{(t+1)}=\arg\max \mathbb{E}(\mathcal{L}(\theta^{(t)},\pi^{(t)} ;X))$$ with ECLL $\mathbb{E}(\mathcal{L}(\theta^{(t)},\pi^{(t)} ;X))$ given by--
\begin{equation*}
\begin{aligned}
&\sum_{i=1}^{n}\sum_{j=1}^{k}\tau_{i,j}^{(t)} \times \log (\pi_j \cdot \mathbb{P}(X=x_i\lvert Z_i=j; \theta))\\
&= \sum_{i=1}^{n}\sum_{j=1}^{k}\tau_{i,j}^{(t)}\log \pi_j + \sum_{i=1}^{n}\sum_{j=1}^{k}\sum_{v=1}^{d} x_{i,d}\tau_{i,j}^{(t)}\log\theta_{j,d}
\end{aligned}
\end{equation*}
yielding estimates $$\pi_j^{(t+1)} = \frac{1}{n}\sum_{i=1}^n \tau_{i,j}^{(t)},\;\theta_{j,v}^{(t+1)} = \sum_{i=1}^n x_{i,v}\tau_{i,j}^{(t)} /d\sum_{i=1}^n \tau_{i,j}^{(t)}.$$
\textbf{Hard clustering} assignments are obtained by calculating
$S_{x_i}=\arg\max_{z}\mathbb{P}(Z_j=z\lvert x_i;\pi,\theta),$
with $S_{{x_i}}\in [K]$ for $i\in [n]$.

As is evident from the model above, we need to decide the number of clusters $K$ before we can fit the above latent factor model. Emphasizing on interpretability of labels along with stability and representativeness of clusters, we use a combination of -\textbf{stability}, \textbf{dominance} and \textbf{interpretability} (as described in the appendix) instead of a naive $WCSS$-criterion to evaluate our clustering performance and decide on the number of clusters.

\section{User personas and insights}
\label{labels}
\section{Personas and insights}
\label{labels}
Having described our methods of constructing personas, we present the summaries of personas based on preferential and behavioral patterns. The significant highlights of these labels are clear characterizations of users in each label. We supplement the labels with interesting insights that can lead to future business actions to understand evolving patterns of both dominant and niche behavioral traits. 
\subsection{User labels for behavioral characterizations}
We give interpretable labels based on the latent structure excavated from the aggregate features described above. See detailed heat-map tables for cluster centers and sizes in the supplement. Below, we list the labels with the cluster sizes reported in percentages (beside the label) and for each label, we give a brief explication of user behavior in that bucket. \\
\textbf{\it Monthly Expenditure}:  Cluster centers represent monthly expense in each of the $13$ price categories (5 rental price categories and 8 purchase categories). The user-labels are-
\begin{itemize}[topsep=0pt,itemsep=-1ex,partopsep=1ex,parsep=1ex]
\item \textbf{Economic Renters} ($71\%$) : $10\$$ in all in a month of activity, with smaller amounts of $2-3\$$ each in the higher renting price categories
\item \textbf{Heavy Renters}  ($21\%$) : $17\$$ in total, mostly around $13\$$ in the $3-5$ rental price category
\item \textbf{Movie Buyers}  ($4.5\%$): $32\$$ in total with one purchase on average in the $16-20$ price category and $1/4$-th of monthly expenses in higher-priced rentals and lower-priced purchases.
\item \textbf{Movie Buffs}  ($2.5\%$): $60\$ $ in all, with around 3 purchases in $10-16$ and around $7$ dollars in $16-20$ price categories.
\end{itemize}
\textbf{\it Frequency of Transaction}:
Cluster centers denote transaction counts in $6$ price ranges, the labels uncovered are--
\begin{itemize}[topsep=0pt,itemsep=-1ex,partopsep=1ex,parsep=1ex]
\item \textbf{Frequent High-End Renters} ($61\%$): over $85\%$ transactions in rentals above $3\$$
\item \textbf{Frequent Low-End Renters} ($21\%$): over $60\%$ and $30\%$ transactions in rentals below and above $3\$$ respectively
\item \textbf{Frequent Movie Buyers \& Sporadic Renters} ($12\%$): $45\%$ transactions in purchases in $8-16\$$ and $35\%$ transactions in rentals as well.
\item \textbf{Frequent Low End Purchasers} ($6\%$): $80\%$ transactions mostly in the $0-8$ price category of purchases
\end{itemize}
\textbf{\it Dominant Genre of Content Consumed}:
The three prime clusters recovered with cluster centers being percentage of monthly transactions in 16 genres are--
\begin{itemize}[topsep=0pt,itemsep=-1ex,partopsep=1ex,parsep=1ex]
\item \textbf{Happy Family} ($23\%$):  content qualifying as family watch with distribution being family genre ($28\%$)--the most consumed genre, followed by animation ($20\%$), comedy ($13\%$); but  no/ almost no crime, horror, romance, thriller
\item \textbf{Drama-Comedy:} ($40\%$) content with dominant genres-- drama ($28\%$), followed by biography ($10\%$), comedy ($10\%$), bit of romance but little or almost nothing as compared to other clusters in terms of consuming family, horror, action,crime
\item \textbf{Action-Horror-Thrill:} ($37\%$) dominant genre is action  ($20\%$), followed by drama  ($15\%$), thriller  ($12\%$), sci-fi  ($8\%$), comedy  ($6\%$), horror  ($5\%$), but little or almost nothing as compared to other clusters in terms of consuming family, comedy-drama, fantasy.
\end{itemize}
\textbf{\it Recency of Content consumed}:  We obtain 3 genre clusters based on the count matrices binned as per release year of content to observe characterizations for recency of content. 
\begin{itemize}[topsep=0pt,itemsep=-1ex,partopsep=1ex,parsep=1ex]
\item \textbf{Latest} ($40\%$): $85\%$ of transactions with release year 2014-15.
\item \textbf{Recent} ($30\%$) : $85\%$ of transactions with release in 2010-13.
\item \textbf{Nostalgic} ($30\%$): About $30\%$ with release in 2000-09 followed by recent and latest content in the remaining $65\%$ of transactions
\end{itemize}
\textbf{\it Time \& Day of Transaction}: Based on habit or preference to transact at a certain time and day of the week, clusters with centers representing counts in each time category of weekday/ weekend are--
\begin{itemize}[topsep=0pt,itemsep=-1ex,partopsep=1ex,parsep=1ex]
\item \textbf{Weekend Evening \& Night} ($24\%$):  $65\%$ of transactions on weekend nights, followed by $25\%$ in evening.
\item \textbf{Weekday Evening \& Night} ($24\%$): $70\%$ of transactions on weekday nights, followed by $20\%$ in evening.
\item \textbf{Weekend \& Weekday Night} ($42\%$): $45\%$ and $35\%$ of transactions on nights of weekdays and weekends
\item \textbf{Weekend Day \& Night:} ($10\%$): $25\%$ of transactions on weekend day time and $60\%$ in weekend nights
\end{itemize}

\subsection{Insights into labels} 
The user personas stated above lend insights that validate common intuitions about consumer behavior and thereby, can render easy motivations for actionable schemes. We describe temporal evolution of user labels - ``how dominant characterizations stay stable while niche ones undergo change". The other attractive property of the discovered latency is stability on a population level. We show a divisive structure in the labels as we increase the number of clusters; this gives us insights into layers within a single characterization. Further, we also use our approach to discover inter-relations between different characterizations as layered clustering instead of a joint one. While the former that discovers layers within a single characterization is not imposed by our algorithm, the latter uses it to discover a layer of characterization within another.\\
\textbf{Temporal nature of labels -\it {``stability of macro characteristics"}:} 
A highlight of the personas is that the clusters uncovered stay stable in terms of size and composition on a population level. This attractive property of consistency allows us to use to model population characterizations consistently across temporal evolution of tenure timelines. At the same time, the personas also also succeed in explaining individual dynamism. That is, migrations do happen on a user level and individual user labels are not static. Our results show these migrations in most categories are never drastic in nature, while we do observe few interesting migrations into far-off labels. These migrations can be explained as \textit{dominant characterizations} reflecting spending capacity, content preferences and habits staying stable over time while \textit{niche characterizations} being more prone to change. 
As specific examples, we see the dominant segment of users transacting in lower-priced categories staying stable in their respective labels over time. However, the niche segment of higher end purchasers keep migrating to lower end categories and migrate back to the niche labels with only availability of new products of their interest. Figure \ref{fig:H2} contrasts the migrations on an individual level between two users, from a dominant and niche spending segment of the population. Another niche segment is a proportion of people who buy content in the \textit{happy family} label; they move to other labels of genre consumption as they also buy content for individual consumption that's different from content consumed with family. The other two labels within genre preference together represent the dominant population and show stability along tenure timelines. \\
\begin{figure}[ht] %
    \centering
    	\includegraphics[height= 3.cm, width=8cm]{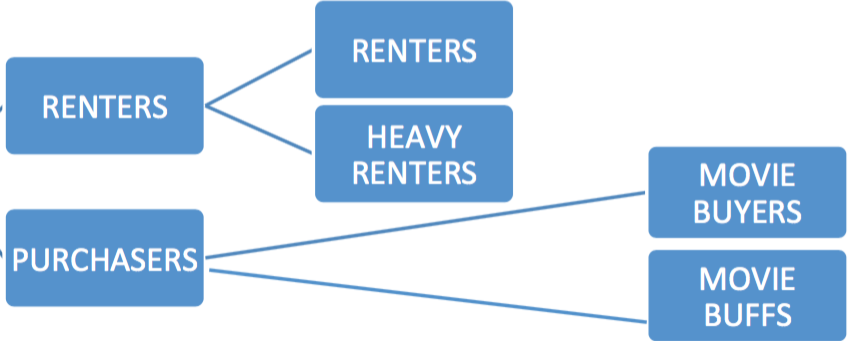}
    \caption{Hierarchy of spend labels: the first two clusters are that of renters and purchasers, which later decompose into the smaller segments within each label. }%
   \label{fig:H2}%
\end{figure}

 
\textbf{Natural hierarchical structure of clusters}   The personas exhibit a natural, divisive, hierarchical structure (not imposed through algorithm), as we increase number of clusters. This lends interesting interpretations on the sub-population of users within broad segments. An example of this is upon clustering users based on monthly expenditure into two clusters, cluster centers represent renters and purchasers, the two main segments of users. Renters break up into economic and heavy renters with 3 clusters. Purchasers mostly decompose into two niche clusters, movie buyers and movie buffs with 4 buckets.\\
Upon clustering count data representing dominant genres consumed by users into two clusters, we see a segment preferring family content over a segment that consumes content not qualifying as family watch. With three clusters, the non-family content consumers decompose into two buckets- one that consumes drama, comedy etc while other prefers thrill inducing content. Figures \ref{fig:H2} and \ref{fig:H3} below give a pictorial depiction of these examples.\\
\begin{figure}[ht] %

    \centering
    	\includegraphics[height= 3cm, width=8cm]{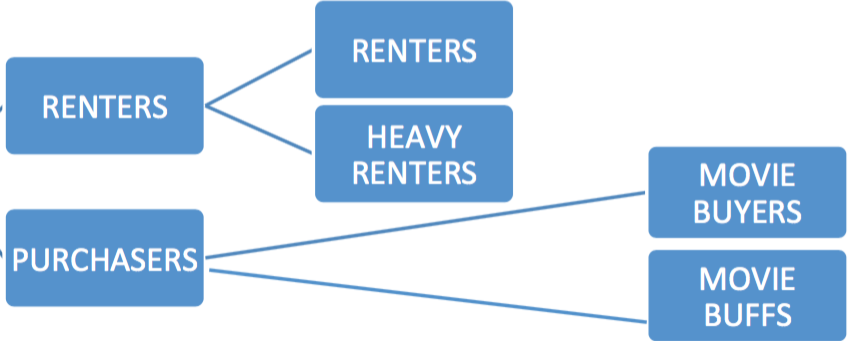}
    \caption{Hierarchy of spend labels: the first two clusters are that of renters and purchasers, which later decompose into the smaller segments within each label. }%
    \label{fig:H2}%
\end{figure}
\begin{figure}[ht] %
    \centering
    	\includegraphics[height= 2.5cm, width=6cm]{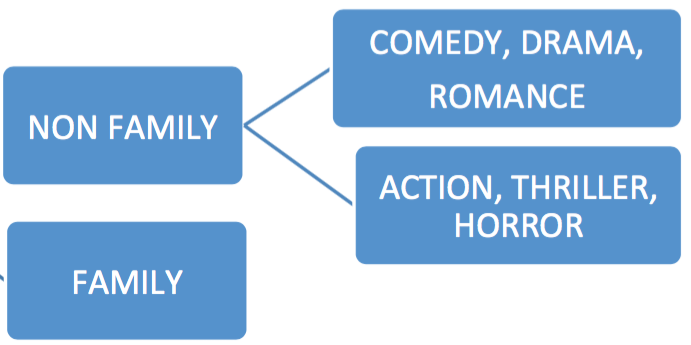}
    \caption{Hierarchy of genre labels: The two major labels upon setting the number of clusters to $2$ is that of the users consuming and not consuming family content. Further clustering splits the consumers that do not transact on family content typically into the two labels: that prefer comedy-drama-romance and action-thriller-horror}%
    \label{fig:H3}%
\end{figure}

\begin{figure}[H] %
\vspace{-1.0em}
    \centering
    	\includegraphics[height= 5cm, width=8cm]{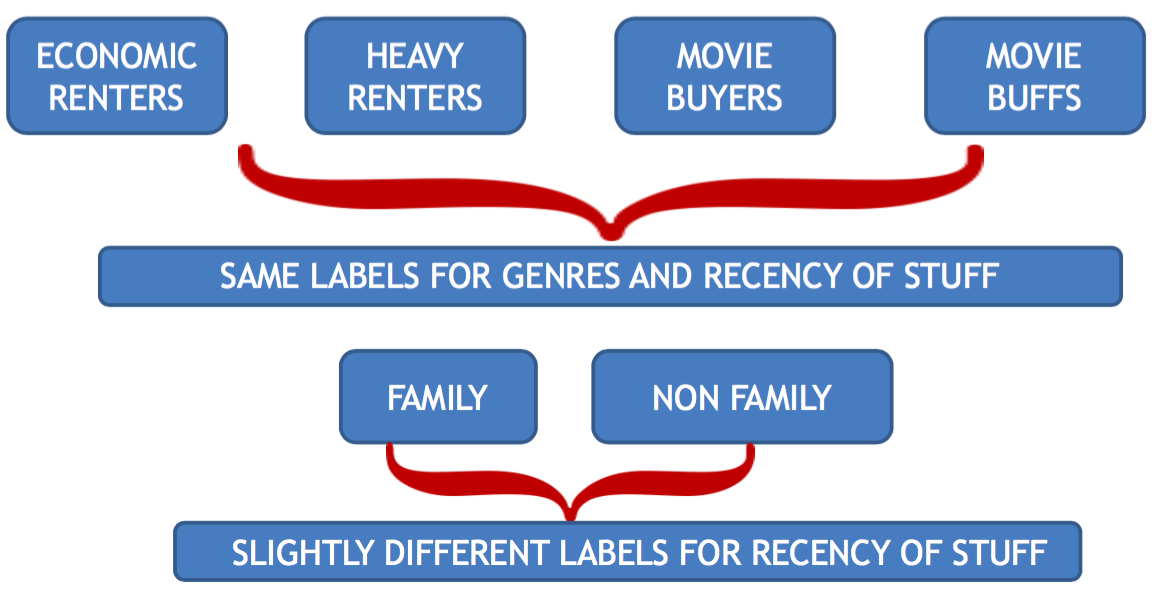}
    \caption{Layers between characterizations: same labels for genre preference within each of the spending labels, but different distributions for recency preference within the genre preference labels}%
    \label{fig:H4}%
\end{figure}

\textbf{Layered structure of clusters}:
We explore inter-relations between characterizations by performing a layered clustering using the mixture model technique. Example being assignment of labels for a characterization like genre preference within clusters for spending behavior. For instance, clustering based on genre preference within the clusters characterizing economic behavior are similar across all economic clusters. Similarly, the clusters for spending behavior are similar across different genres. This observation statistically validates that genre preferences of consumers is independent of their economic budget. A similar observation goes for recency and economic behavior. On the other hand, we see different clustering results for recency of content when clustered within the genre clusters, with the category preferring family content showing more inclination towards more classic content that the other drama-based or thrill inducing categories that prefer more recent content. 
Figure \ref{fig:H4} explores inter-relations between different characterizations, specifically that genre preferences stay the same across all spending labels. This is in contrast to users that are inclined to buy content not qualifying as family watch, who prefer more recent content.

\section{Integration of personas in personalization}
We demonstrate utility of personas through an effective integration of information from personas into personalization. Specifically, we focus on a CTR predictive model where the goal is to predict $p_{u,i}$, the probability that an user $u$ transacts on an item $i$. The scope of utilizing persona information extends to other popular models in collaborative filtering. We conclude the paper by discussing such possibilities, where one can integrate personas into other commonly used models and expect to attain a relevance-scalability-interpretability tradeoff.
\subsection{CTR: relevance-scalability-interpretability balance}
We model the CTR problem to predict transactional probabilities through an $\ell_1$ penalized logistic regression model that is trained per item. Such a finer-grained model at the item level captures the item specific interest in users, leading to more accurate predictions in \cite{zhang2016glmix}. The challenge in such models is however, the sparsity of the transactional database, with about $\bf 1\%$ users transacting on any given item. To overcome this imbalance and avoid bias towards the outcome of \textit{not transacting at all}, we sub-sample for every positive sample (users who transacted) 5 negative samples (users who did not transact). The gain with summarized information from personas can be described as a balance between scalability of the training model, interpretability of feature space and relevance of predictions. \\
\textbf{Relevance}-deliver relevant recommendations to users, quantified by the quality of prediction in transactional probabilities. To fix notations, we denote the 
the evaluation metric to assess the performance of the predictive model as $\mathcal{F}$ on a test set. With the training model $M^*$ giving predicted labels $\widehat{\text{label}}_{M^*}$, the predictive ability is given by $\mathcal{F}(\widehat{\text{label}}_{M^*},\text{label}_{\text{test}})$. $\mathcal{F}$ here, is the mean AUC over the $\bf 100$ most popular items in the content catalogue.\\
\textbf{Scalability}-reduce size of input feature and sample space (leads to reduction in regression size) by using lower dimensional persona features. Information from personas can be encoded as soft clustering features or incorporated as hard clusters via a model trained per cluster. This brings significant reduction in regression dimensions which in turn, facilitates storage and future use of these feature vectors in the same or other predictive models.\\
\textbf{Interpretability}-retain meaning to feature space as opposed to random lower dimensional projections which seldom lend business insights. With a meaningful feature set, we can reutilize the same features in a host of predictive tasks and use them in easy debugging of models. While relevance and scalability can be quantified, there is no measure of interpretability and is open to subjective assessment.

\begin{figure}
    \centering
    	\includegraphics[height= 5cm, width=6.5cm]{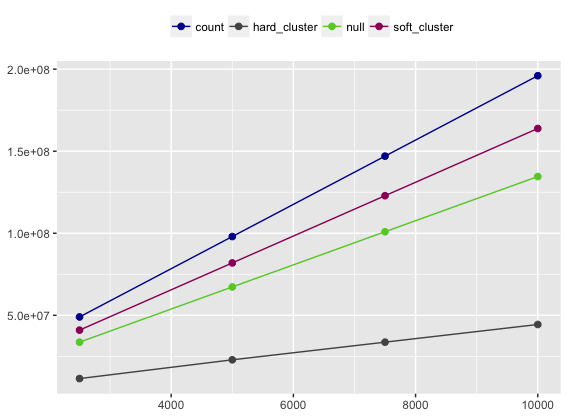}
    \caption{Computational complexity $\mathcal{O}$ as sample size per item $n_i$ varies and $n_{i,c} = [n_i/3]$ varies. The null model can be seen to more scalable than the model based on counts or soft-clustering features, but is seen to lose $13\%$ power in comparison to the baseline model based on counts. The persona based models are comparable to the baseline in terms of predictive ability, but are more scalable. The gain with hard cluster membership based model is clearly considerable.}%
    \label{fig:cc}%
\end{figure}

The trade-off in the above criteria arises as we can use a baseline model with the count features that were used to recover latent user labels as regressors. However, there is a significant computational cost associated with a higher regression size of the baseline based on these aggregate features, without using any knowledge of personas. We see a clear reduction in regression size and the associated complexity with integration of persona information at the cost of losing only a mere $2\%$ predictive ability in Figure \ref{fig:cc}. Scalability of regression size with comparable predictive power as the baseline model alongside retaining clear meaning of feature space is the trade-off achieved in CTR prediction with persona information.
The take away is that persona features can be used to construct interpretable, lower dimensional regressors that preserve predictive power. An added advantage of incorporating these summary features in a model with sub-sampled users is preservation of privacy of individual users and also, of individual transactions in using summaries over a random set of users. 

To describe our model and results, we use $X_u$ to denote the feature vector corresponding to user $u$. This feature vector can be based on 3 characterizations: ME (monthly expenses), DG (dominant genre), CR (content recency). Information from personas can be incorporated into $X_u$ in different ways, yielding different models. In particular, we construct feature vectors using the personas on ME, DG and CR in the following forms - denoted by \textbf{(c), (s), (h)} and \textbf{(-)} respectively. \textbf{(c)} is used in the baseline model with count features based on a particular characterization, \textbf{(s)} and \textbf{(h)} integrate soft and hard clustering information based on characterizations. \textbf{(-)} uses neither count nor persona information, we call this the null model. These are summarized below:

\textbf{(c)} a feature vector with distribution of ME in price categories and/ or count vectors for DG/CR in feature bins (directly using the constructed features) : uses aggregate features, but no additional knowledge from latent personas\\

\textbf{(s)} a feature vector of soft clustering values in the form of distances of count features from their respective cluster centers\\

\textbf{(h)} incorporates hard clustering information for a characterization by training a model cluster-wise\\

\textbf{(-)} does not include any information from a characterization at all.\\

\begin{figure}
    \centering
    	\includegraphics[height= 4cm, width=7.5cm]{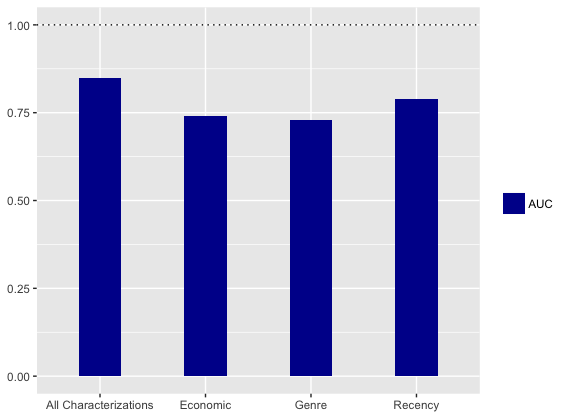}
    \caption{AUC of models based on $X_u$ that uses different characterizations: Recency of a content item is the most predictive of the consumer's interest in transaction.}%
    \label{fig:rel}%
\end{figure}

Below, we describe the different CTR models and discuss results on the three criteria trade-off. \\
We achieve a \textbf{gain in relevance} with information from each added characterization, either in the form of soft clustering/ hard clustering/ count feature. Figure \ref{fig:rel} highlights the relevance of each characterization in the CTR model. CR (recency) is seen to the most informative characterization adding the most to AUC. 
Denote $n_i$ as the samples per item and $n_{i,c}$ as samples per item, per cluster, $p$ the number of predictive features, $\mathcal{O}$ as the complexity of regularized logistic with sample size and regression dimension. The below table compares different techniques illustrating how our proposed integration of user personas into personalized recommendation achieves a tradeoff between relevance and scalability (noting that interpretability comes alongside using summary information from personas). 

The baseline model in the first row of the table; representing the model with all count features \textbf{(c)}. We see a significant reduction in the predictive power when we do not incorporate any information from the recency feature, this is depicted by the fourth row of the table. When we train a model per recency cluster using \textbf{(h)}, we lose $1\%$ of predictive power, but reduce the sample size for the training model on each cluster with a reduction in the sample space. We see a similar predictive power when we use soft clustering recency feature \textbf{(s)}, but a significant reduction in the size of feature space. While we do not incorporate all $64$ combinations of $\textbf{(c)},\textbf{(s)}, \textbf{(h)},\textbf{(-)}$, we see that using soft clustering features for all the $3$ characterizations leads to a loss of only $2\%$ AUC. This is represented in the last row of below table. The computational gain, however, is seen to be significant even in a simple regression model that scales in complexity as $p^2$ with the size of feature space. Figure \ref{fig:cc} shows this for $n_i$ and $n_{i,c} = [n/3]$ vary per item. Interpretability is inherent in these models due to the clear meaning of soft clustering features that represent distances from cluster centers or hard-coded cluster memberships in training a model based on similar users. 
\begin{center}
\captionof{table}{Tradeoff in AUC versus complexity based on size of sample and feature space}
     \scalebox{0.8}{\begin{tabular}{||c c c c c c c||}
 \hline
 Recency & Genre & Economic & $\mathcal{F}$ & n & p & $\mathcal{O}$ \\ [0.5ex]
 \hline\hline
c & c &  c &  0.85 & $n_i$ & $140$ & $O\left(n_i \times 140^2\right)$ \\  [1ex]
 \hline
 s & c & c & 0.84 & $n_i$ & $128$ & $O(n_i \times 128^2)$\\ [1ex]
  \hline
 h & c & c & 0.84 & $n_{i,c}$ & $116$ & $O(n_{i,c} \times 116^2)$\\ [1ex]
  \hline
 - & c & c & 0.75 & $n_i$ & $116$ & $O(n_i \times 116^2)$\\ [1ex]
  \hline
   s & s & s & 0.82 & $n_i$ & $40$ & $O(n_{i} \times 40^2)$\\ [1ex]
  \hline
\end{tabular}}
\end{center}
\subsection{Persona based collaborative filtering models} 
We finally discuss few models based on popular collaborative filtering techniques that can incorporate information from personas to retain predictive power while gaining in scalability for practical implementations.\\
\emph{\bf User based nearest neighbor similarity:} This approach is based on a similarity metric $\text{sim}(u,v)$ (examples include Jaccard, cosine etc.) to predict a weighted average rating based on similarity between users who transacted on the same items. Denoting $U(i)$ as the set of users who transacted on the same item $i$, amount of money user $u$ is willing to spend on item $i$ can be predicted as
\[r_{u,i}={\sum_{v\in U(i)}\text{sim}(u,v) r_{v,i}}/{\sum_{v\in U(i)}\text{sim}(u,v)}\]
and probability that user $u$ transacts on item $i$ as
\[p_{u,i}={\sum_{v\in U(i)}\text{sim}(u,v)}/{\sum_{v}\text{sim}(u,v)},\]
with $1_{v,i}$ as indicator if user $v$ transacted on item $i$. Similarity approaches have scaling issues with high computational cost associated with searching through set of users or even the top $K$ similar users in the set $U(i)$. Persona information can bring in gain in prediction accuracy, also offering better scalability via limiting search of top $K$ neighbors to already formed personas. 

We could use clusters from most representative time point of activity for predictions. Alternately, we can use temporal persona information for prediction with the scope of leveraging differently on time points through a weighted similarity prediction along the tenure timeline. Denoting time points of transaction history (months of tenure timeline) as $t$ with weights $w_t$ (that can be tuned) and features-$u_t$ for user $u$, $U(t,i)$ as the set of users who transacted on the same item $i$  and $C(t,u)$ the set of users present in the same cluster as user $u$ at time $t$, ratings at a time point $T$ leveraging on temporal history till time $T$ can be predicted as
\begin{center}
\captionof{table}{Ratings in CF: Clustering buckets $C(u)$}
     \begin{tabular}{ | l | l | l | l | l | }
     \hline
     $r_{u,i}(T)=\cfrac{\sum_{t\leq T}\sum_{v\in U(t,i)\cap C(t,i)}w_t\text{sim}(u_t,v_t) r_{v,i}}{\sum_{t\leq T}\sum_{v\in U(t,i)\cap C(t,i)}w_t\text{sim}(u_t,v_t)}$\\ \hline
     $p_{u,i}(T)=\cfrac{\sum_{t\leq T}\sum_{v\in U(t,i)\cap C(t,i)}w_t\text{sim}(u_t,v_t) r_{v,i}}{\sum_{t\leq T}\sum_{v\in C(t,i)}w_t\text{sim}(u_t,v_t)}$ \\ \hline
     \end{tabular}
\end{center}
\emph{\bf Latent factor model:} Without clustering information, the vanilla model with latent factors $q_i$ for item $i$ and $p_u$ for user $u$  is
\[\hat{r}_{u,i}=\mu+b_i+b_u +q_i^T p_u,\;\]
 solved either through stochastic gradient descent or alternating least squares. Letting $\mathcal{A}$ to be a set of attributes and $a$ a cluster for $\mathcal{A}$, persona information can be incorporated into the above model by (see below table for ratings in each case in that order)--
\begin{enumerate}[topsep=0pt,itemsep=-1ex,partopsep=1ex,parsep=1ex]
\item adjusting for biases per cluster
\item enhancing user representation in the form of latent factors for cluster memberships learnt with $y_a \in \mathcal{A}$--a latent factor for each cluster $a$ in set of characterizations (see \cite{koren2009matrix})
\item hard wiring clustering information as features in the form of an enhanced user feature with a latent component $p_u$ concatenated with known added features $\tilde{p}_{u}$
\item training latent factor model per cluster with $c_a$ being clusters corresponding to some attribute $a$.
\end{enumerate}
\begin{center}
\captionof{table}{Ratings in CF: Adding Clustering Info to Vanilla}
     \begin{tabular}{ | l | l | l | l | l | }
     \hline
     (a) & $\hat{r}_{u,i}=\mu+ b_i + b_u +\sum_{a\in \mathcal{A}(u)}b_a + q_i^T p_u$ \\ \hline
     (b) & $\hat{r}_{u,i}=\mu+ b_i + b_u + q_i^T (p_u +\sum_{a\in \mathcal{A}(u)}y_a)$ \\ \hline
     (c) &  $\hat{r}_{u,i}=\mu+ b_i + b_u + \tilde{q}_i^T (p_u: \tilde{p}_u)$ \\ \hline
     (d) & $\hat{r}_{u,i}^{c_a}=\mu_{c_a}+b_i+\mathbb{I}_{u\in c_a}b_u + \mathbb{I}_{u\in c_a}q_i^T p_u$ \\ \hline
     \end{tabular}
\end{center}

\section{Concluding remarks}
The work offers temporally evolving personas that lend new perspectives and actionable insights into behavioral patterns of VoD users as they age in the system. As highlighted, our personas do possess the cluster stability on a macro level, while being able to represent dynamic niche characterizations at the same time. Our mixture approach together with the choices of granularity and timeline of comparison and the engineered features give rise to a consistent and robust latent model. That is. insights derived from a study of personas at any time point are also likely to apply to future clusters and models built using these clusters. Information from efficiently built personas can achieve a much practical and vital \textit{relevance-scalability-interpretability} tradeoff in recommendations, highlighted in the work with predictive models that are trained and tested on VoD data. An untapped area of application is churn analysis, see \cite{kapoor2014hazard}, aiming to improve user retention and interest. One can create user buckets based on longevity in system or use existing personas to predict when users slip into a state of coma in the system. A potential future direction also includes a possible tradeoff between privacy and predictive power in models based on persona features. Finally, the methods, guarantees and perspectives from this work can be extended to other domains of personalization and realized in a host of other predictive tasks.

\bibliographystyle{plainnat}
{\large{\bibliography{sigproc}}}

\section{Appendix}
\subsection{Evaluation of clustering performance}
The clusters that define user personas in this work satisfy the below three criteria--\textbf{stability}, \textbf{dominance} and \textbf{interpretability} as defined below.
We define these below:\\
\emph{\bf $\epsilon-\delta$ Stable clusters:}
Theoretically, clustering stability is achieved if similar results are obtained when applied to several data sets from the same underlying model or simulated by the same data generating process. In our case, clusters are defined to be $\epsilon-\delta$ stable on a population level if the cluster centers of the most similar clusters are $\epsilon$ close in $\ell_2$ norm and the corresponding cluster sizes vary by a fraction of at most $\delta$ when the mixture approach is repeatedly applied to random sub-samples of size $50\%$ (without replacement) of the original data set.\\
\emph{\bf $K_{\text{max}}-\kappa$-thresholded Dominant clusters:}
Typically, clustering is said to be dominant if clusters obtained are representative of large segments and also, of niche segments of the population. A precise way of obtaining such clusters which capture both the dominant and niche characterizations of a population in our case is by setting the cluster sizes to be above a threshold. In our case, clusters are said to be $K_{\text{max}}-\kappa$-thresholded dominant clusters if cluster sizes are at least greater equal to $\kappa>0$ and number of clusters is upper bounded by $K_{\text{max}}$.\\
\emph{\bf Interpretable clusters:}
Interpretability is hard to be made precise mathematically. We define it loosely in terms of clear meanings to clusters that can help characterize customers distinctly.
Clusters are deemed interpretable if the summarized information from the clusters can be lent to intuitive explanation and can lead to actionable insights that can help improve key business performance indicators.\\

A combination of these criteria helps us determine the number of clusters in constructing user personas.

\subsection{Details on user labels: composition}
Below, we give details of clusters of the characterizations: {\it Monthly Expenditure}, {\it Frequency of Transaction}, {\it Dominant Genre of Content Consumed}, {\it Recency of Content consumed}, {\it Time \& Day of Transaction}. The rows of the heat tables represent the clusters with the categories mentioned at the top of the tables, in the same order as enlisted in user labels in the main draft. For {\it Monthly Expenditure}, the cluster centers denote the average monthly expenses in each price category. For all other characterizations, the cluster centers show the percentage of total monthly transactions in that particular feature bin.
\medskip

The below tables shows the average monthly expenditure in the $5$ rental price categories and the $8$ purchase price categories respectively for the ME labels with rows in the order of \textit{Economic Renters, Heavy Renters, Movie Buyers, Movie Buffs}. \\

\centering{\pgfplotstabletypeset[
    color cells={min=-2.05,max=25.00},
    col sep=comma,
    /pgfplots/colormap={whiteblue}{rgb255(0cm)=(255,255,255); rgb255(1cm)=(0,0,188)},
]{
    0, 0-1, 1-3, 3-5, $>5$
    0,  0, 0.97, 2.38, 2.41
    0, 0, 1.48, 13.02, 1.79
    0, 0,  0.82, 3.35, 2.46
   0, 0, 1.8, 6.02,3.19
    }}
\medskip

\centering{\pgfplotstabletypeset[
    color cells={min=-2.05,max=25.00},
    col sep=comma,
    /pgfplots/colormap={whiteblue}{rgb255(0cm)=(255,255,255); rgb255(1cm)=(0,0,188)},
]{
    0, 0-3,  3-5, 5-8, 8-10, 10-16, 16-20, $> 20$
    0, 0.56, 0.01, 0.04, 0.33, 1.07, 0, 0.44
   0, 0.36, 0.01, 0.03, 0.34, 0.87, 0.04, 0.16
    0, 0.56, 0.02, 0.10, 1.10, 3.74, 23.95, 0.09
     0, 1.11, 0.08, 0.39, 4.29, 39.86, 6.49, 2.12
    }
}

\bigskip

The next heat map gives the compositions of \textit{Frequent High-End Renters, Frequent Low-End Renters, Frequent Movie Buyers \& Sporadic Renters, Frequent Low End Purchasers} under the characterization describing frequency of transaction . Each cell represents the percentage of monthly transactions corresponding to a rental (R) or purchase (P) price category.\\
\medskip

\centering{\pgfplotstabletypeset[
    color cells={min=-1,max=100},
    col sep=comma,
    /pgfplots/colormap={whiteblue}{rgb255(0cm)=(255,255,255); rgb255(1cm)=(0,0,188)},
]{
    R: 0-3, R: $>3$, P: 0-8, P: 8-16,  P:16-20,  P: $>20$
    10, 86, 0, 3,1, 0
  60, 33, 2, 5, 0, 0
     8, 29, 4, 43, 13, 3
     5, 12, 78, 4, 0, 1
    }}
    
\bigskip

The below heat-map gives the compositions of the labels- \textit{Happy Family, Drama-Comedy, Action-Horror-Thrill}, the three dominant clusters that represent user's preference for genres of content.\\

\medskip

\centering{\pgfplotstabletypeset[
    color cells={min=-1,max=100},
    col sep=comma,
    /pgfplots/colormap={whiteblue}{rgb255(0cm)=(255,255,255); rgb255(1cm)=(0,0,188)},
]{
    Drama, Comedy, Action, Family, Ani, Thrill, Biog, Scifi
    5, 13, 5 , 28, 20, 2, 2, 2.5
    28, 10, 4, 1, 0, 3, 10, 0
    15, 6, 20, 1, 0, 12, 2, 8
    }}

\centering{\pgfplotstabletypeset[
    color cells={min=-1,max=100},
    col sep=comma,
    /pgfplots/colormap={whiteblue}{rgb255(0cm)=(255,255,255); rgb255(1cm)=(0,0,188)},
]{
    Crime, Sup.her, Com-dr, Fan, Horr, Rom, Kids, Misc.
    0, 4, 2, 4, 0, 0, 3, 7
    4, 0, 5, 7, 0, 4, 0, 20
    4, 4, 1, 1, 5, 2, 0, 14
    }}

\bigskip
\bigskip

The next heat map gives the cluster centers for recency labels in the order of \textit{Latest, Recent, Nostalgic}.
\medskip

\centering{\pgfplotstabletypeset[
    color cells={min=-1,max=100},
    col sep=comma,
    /pgfplots/colormap={whiteblue}{rgb255(0cm)=(255,255,255); rgb255(1cm)=(0,0,188)},
]{
    older than 1990, 1990-2000, 2000-10, 2010-14, 2014-15
    0, 0, 3, 9, 88
     0, 0, 3, 88, 9
    8, 9, 28, 32, 33    
    }}

\bigskip

Finally, below are cluster centers for user labels that characterize transactional habits in terms of time and days of transaction. The rows of the heat-map table are again in the order
of \textit{Weekend Evening \& Night, Weekday Evening \& Night, Weekend \& Weekday Night, Weekend Day \& Night.} The first table gives the percentage of monthly transactions done on weekday slots and the second table gives the same on weekend slots.\\
\medskip

\centering{\pgfplotstabletypeset[
    color cells={min=-1,max=100},
    col sep=comma,
    /pgfplots/colormap={whiteblue}{rgb255(0cm)=(255,255,255); rgb255(1cm)=(0,0,188)},
]{
    Wk 10-5,  Wk 5-10, Wk 10$-$5
    0, 1, 10 
    8, 18, 68
    4, 9, 45      
     2, 2, 10
    }}
   
   \medskip 
\centering{\pgfplotstabletypeset[
    color cells={min=-1,max=100},
    col sep=comma,
    /pgfplots/colormap={whiteblue}{rgb255(0cm)=(255,255,255); rgb255(1cm)=(0,0,188)},
]{
    Wd 10-5, Wd 5-10, Wd 10$-$5
   1, 24, 62
     0, 0, 5 
    3, 6, 31    
    4, 7, 56
    }}

\end{document}